\documentclass[letterpaper]{article}
\usepackage{aaai}
\usepackage{times}
\usepackage{helvet}
\usepackage{courier}
\usepackage{latexsym}
\usepackage{multirow}
\usepackage{graphicx}
\usepackage{amsmath}
\usepackage{amssymb}
\usepackage{amsfonts}

\DeclareMathOperator*{\argmax}{argmax}

\newcommand{\namecite}[1]{\citeauthor{#1}~\shortcite{#1}}

\begin{document}
%
\title{Beyond Word-based Language Model in Statistical Machine Translation}
\author{Jiajun Zhang$^{\dagger}$, Shujie Liu$^{\ddagger}$, Mu Li$^{\ddagger}$, Ming Zhou$^{\ddagger}$ and  Chengqing Zong$^{\dagger}$ \\
	$^{\dagger}$National Laboratory of Pattern Recognition, CASIA, Beijing, P.R. China \\
	{\small \{\tt jjzhang,cqzong\}@nlpr.ia.ac.cn} \\
	$^{\ddagger}$Microsoft Research Asia, Beijing, P.R. China \\
	{\small \{\tt shujliu,muli,mingzhou\}@microsoft.com}}

\maketitle
\begin{abstract}
\begin{quote}
Language model is one of the most important modules in statistical machine translation and currently the word-based language model dominants this community.
However, many translation models (e.g. phrase-based models) generate the target language sentences by rendering and compositing the phrases rather than the words. Thus, it is much more reasonable to model dependency between phrases, but few research work succeed in solving this problem.
In this paper, we tackle this problem by designing a novel phrase-based language model which attempts to solve three key sub-problems: 1, how to define a phrase in language model; 2, how to determine the phrase boundary in the large-scale monolingual data in order to enlarge the training set; 3, how to alleviate the data sparsity problem due to the huge vocabulary size of phrases.
By carefully handling these issues, the extensive experiments on Chinese-to-English translation show that our phrase-based language model can significantly improve the translation quality by up to +1.47 absolute BLEU score.
\end{quote}
\end{abstract}

\section{Introduction}
As one of the most important modules in statistical machine translation (SMT), language model measures whether one translation hypothesis is more grammatically correct than other hypotheses.
Since the beginning of the statistical era for machine translation, word-based language model dominates this community. When the word-based SMT was first proposed, the model generates the target translation word by word and we need to calculate how fluency is the concatenation of the individual words. Thus, the word-based language model becomes a natural choice, see Figure 1(b) for illustration.
However, the more sophisticated translation models (e.g. the phrase-based models \cite{koehn2007moses,blackwood2009large}) manipulate the phrases rather than the words during the decoding stage. The phrases come from natural language texts and it is less necessary to model the dependency among the inner words of the phrases. Instead, it is much more beneficial to model the probability distribution of a new target phrase conditioned on the previously generated target phrases, see Figure 1(c) for demonstration. Obviously, the word-based language model cannot depict the dependency between the phrases.

Although it is very promising if we can design a good phrase-based language model, few research work succeed in solving this problem in statistical machine translation. In automatic speech recognition community, \namecite{heeman1997deriving} attempt to derive a phrase-based language model. However, their method estimates the conditional probability of the phrases by backing off to words rather than considering the phrases as the inseparable units.  \namecite{baisa2011chunk} first proposed the chunk-based language model (including phrase-based) in machine translation but did not give a solution. Recently, \namecite{xu2015phrase} designed a direct algorithm for phrase-based language model in statistical machine translation. In their method, phrase can be any word sequence. The phrase vocabulary is huge and the data sparsity problem is very serious. It leads to difficulty in probability estimation for phrase-based language model.

\begin{figure}
\centering
\includegraphics[scale=0.48]{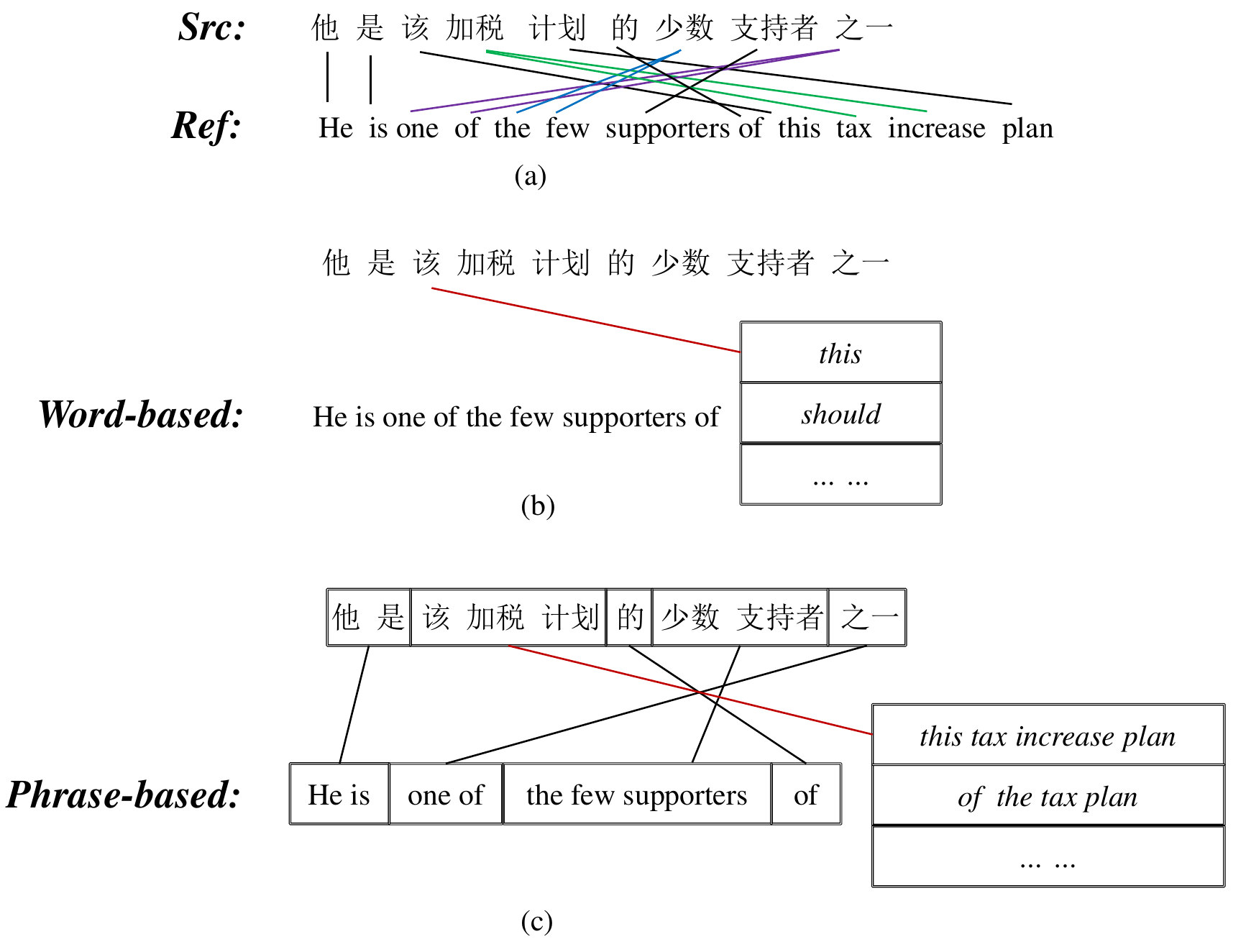}
\caption{An example comparing the word-based translation and the phrase-based translation. (a) shows the Chinese source sentence and English reference. (b) illustrates a decoding phase of the word-based model translating the third Chinese word. (c) demonstrates a decoding stage of the phrase-based model translating the second Chinese phrase.}
\label{Fig.1}
\end{figure}

Why so few researchers succeed in proposing good solutions to the phrase-based language model? We believe the reason lies in three difficulties: 1, the phrase can be any word string, and it is hard to give a good definition for the phrases; 2, it is a problem how to transform the word-based large-scale monolingual data into the phrase-based corpus for phrase-based language model training; 3, due to the huge size of the phrase vocabulary, it is a question how to alleviate the data sparsity problem.

In this paper, we aim at tackling the above three problems and propose an effective phrase-based language model. The key idea behind is that, through the word-aligned parallel sentences, we define and obtain the minimal phrases. Then we adopt a sequence labelling approach to find the minimal phrase boundary for the large-scale monolingual data. Finally, the deep neural network (DNN) is leveraged to investigate the data sparsity problem of the phrase-based language model. We make the following contributions:

\begin{itemize}
\item In order to make the granularity of the phrase stable enough, we define the minimal phrases inspired by the concepts of minimal translation units (MTU) in \cite{zhang2013beyond}. We further regard the phrase boundary recognition in large-scale monolingual data as a sequence labelling task.

\item We investigate the data sparsity problem of the phrase-based language model by introducing a deep neural network. We show DNN has the potential to alleviate the data sparsity problem although it is not good enough currently due to the strict constraint of vocabulary size.

\item Our phrase-based language model achieves substantial improvements on the large-scale translation tasks.
\end{itemize}

\section{Phrase-based Language Model}
For the translation models which generate the output by rendering and compositing the target language phrases, no matter whether the translation is obtained using the left-to-right algorithm \cite{koehn2007moses} or the bottom-up approach \cite{xiong2006maxent}, the phrase-based language model is proposed to measure whether one output phrase sequence is more grammatically correct than others.

Given a partial translation candidate in the form of phrase sequence $t=t_{p0}t_{p1}...t_{pn}$, the phrase language model attempts to calculate the following probability:

\begin{equation}
\begin{split}
P(t) &= P(t_{p0}t_{p1}...t_{pn}) \\
& = p(t_{p0})p(t_{p1}|t_{p0})...p(t_{pn}|t_{p0}...t_{p(n-1)})
\end{split}
\end{equation}

Taking the phrase-based decoding in Figure 1(c) for example, the translation candidate in the form of phrase sequence is {\em He is \$ one of \$ the few supporters \$ of \$ this tax increase plan}, in which {\$} denotes the phrase boundary. 
The first question is that can we just calculate Equation 1 by setting $t_{p0}=$ {\em He is}, $t_{p1}=$ {\em one of}, $t_{p2}=$ {\em the few supporters}, $t_{p3}=$ {\em of} and $t_{p4}=$ {\em this tax increase plan}? In practice, it is not very reasonable due to three reasons.

First, this kind of phrases lacks of a well-formed definition and the training data for the phrase language model is hard to construct. In order to have a better understanding, let us first look at the way generating these phrases. The target phrases used in the decoding stage are all from the phrasal translation rules, and the phrasal translation rules are extracted from the word-aligned parallel sentences. Any word sequence pair satisfying the word alignment becomes a phrasal translation rule. If the sentence pair in Figure 1(a) is used as a training instance and at most three source-side words are allowed, we can extract 17 phrasal translation rules as shown in Figure 2. We see that four target English phrases containing the word {\em tax}. That is to say, there are at least four ways partitioning the phrases containing the word {\em tax}. Thus, which partition should be adopted to train the phrase-based language model? Obviously, lacking of a well-defined phrase concept makes the construction of the training data impossible.

Second, during decoding the phrase-based SMT considers multiple overlapping segmentations of the same sentence. Thus, one problem with the phrase-based language model is that a unique segmentation of a sentence into phrases is not available a-priori.

Third, the phrase vocabulary is too huge to accurately estimate the parameters. The bilingual training data consisting of millions of sentence pairs may extract distinct target phrases in tens of or hundreds of millions. Consequently, it is impossible for us to collect enough training data for accurate parameter estimation.

\begin{figure}
\centering
\includegraphics[scale=0.5]{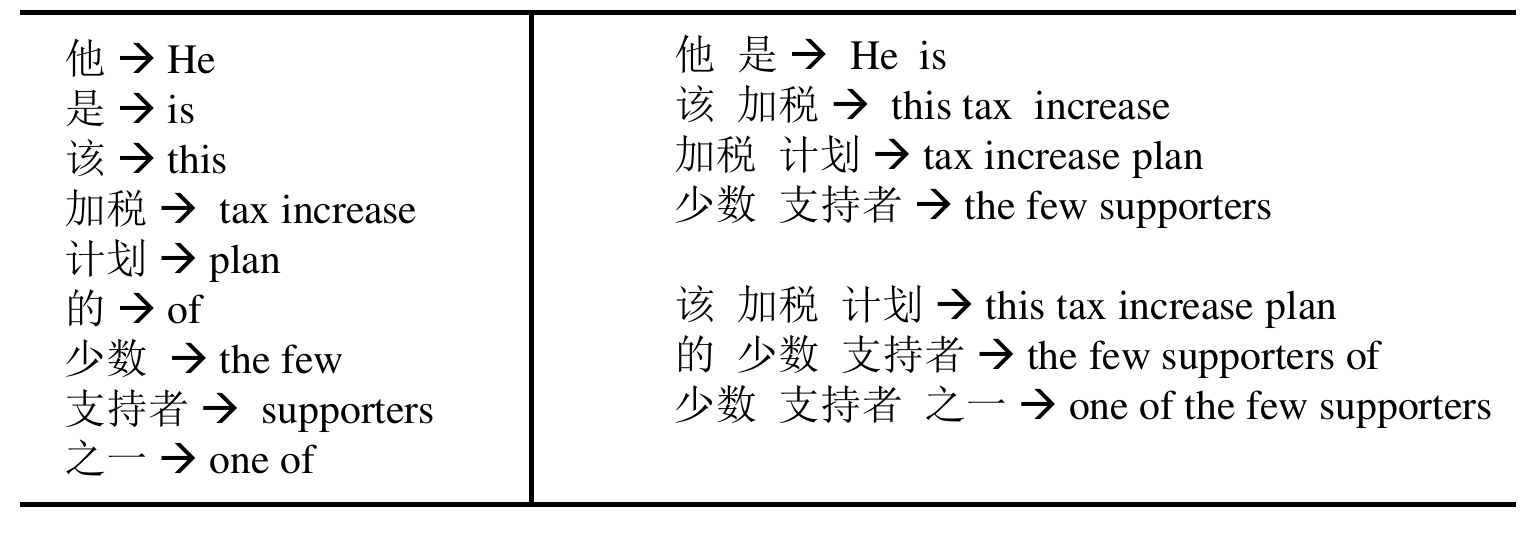}
\caption{The phrasal translation rules extracted using Figure 1(a) as a training instance.}
\label{Fig.2}
\end{figure}

To make it possible for training data construction and accurate parameter estimation, we propose to define the concept {\textbf{\em minimal phrase}}.

\subsection{Definition of Minimal Phrase}
We define the minimal phrase in the context of the bilingual sentence pairs. Informally, a minimal phrase in the target language sentence is a continuous word sequence which contains the minimum words without violating the constraints of word alignment.

Given a word-aligned parallel sentence pair $(s,t,A)$, in which $s=s_0s_1...s_{m-1}$ is the source language sentence, $t=t_0t_1...t_{n-1}$ is the target language sentence and $A \subseteq \{0..m-1\} \times \{0..n-1\}$ is the word alignment between the source and target words. A minimal phrase is a target word sequence $t_i..t_j$ which satisfies the following constraints:

\begin{itemize}
\item There exists a source word sequence $s_k..s_l$ (may be empty) such that for all aligned pairs $(i',k')$, we require $i \le i' \le j$ if $k \le k' \le l$.
\item There is no smaller non-empty target word sequence $t_a...t_b$ ($i < a \le b \le j$ or $i \le a \le b < j$) which meets the first condition.
\end{itemize}

The first constraint is identical to that of the phrasal translation rule extraction \cite{koehn2007moses} except that the source word sequence can be empty in our case. The second constraint guarantees that the phrase is the shortest. From the view of the phrase pair, it is similar to that of minimum translation units \cite{zhang2013beyond}. However, there are two main differences: 1, we only care about the minimal phrase in the target language side; 2, the target phrase in minimum translation units can be empty \cite{zhang2013beyond} while our target minimal phrase must contain one word at least.

\begin{figure}
\centering
\includegraphics[scale=0.4]{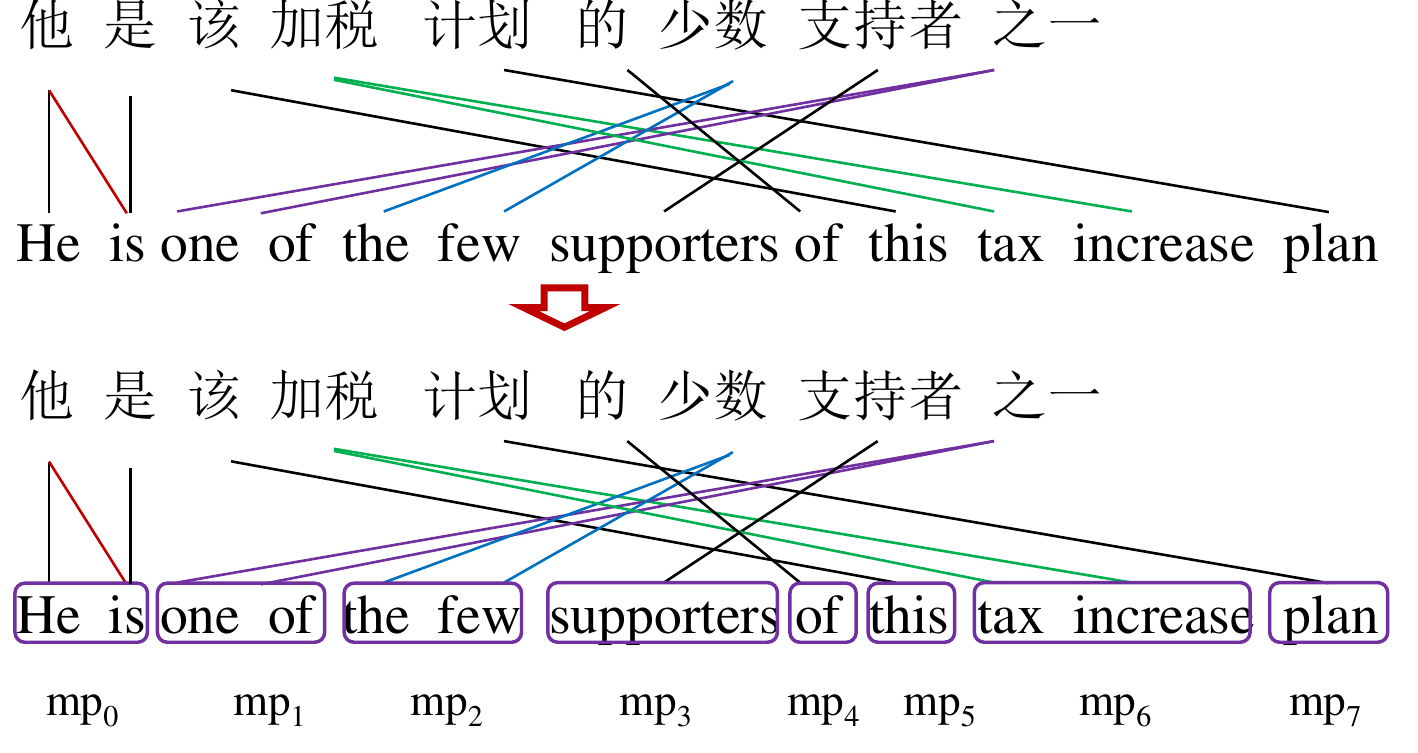}
\caption{An illustration for minimal phrases in target sentence. This automatic word alignment is slightly different from the manually labelled word alignment in Figure 1(a).}
\label{Fig.3}
\end{figure}

All the identified minimal phrases in the target language sentence will form a partition for this sentence. Among all the partitions that satisfy the word alignment constraints, the set of the minimal phrases is the partition with the shortest average phrase length. It should be noted that for each word-aligned sentence pair, the partition of the minimal phrases is \textbf{unique}. Moreover, \textbf{any} target phrase of the phrasal translation rules can be viewed as the composition of the minimal phrases. For example, Figure 3 shows a word-aligned sentence pair and the its corresponding minimal phrases. All the five target phrases used in Figure 1(c) can be obtained by concatenating the minimal phrases. $t_{p0}=mp_0$, $t_{p1}=mp_1$, $t_{p2}=mp_2+mp_3$, $t_{p3}=mp_4$ and $t_{p4}=mp_5+mp_6+mp_7$. Then, Equation 1 becomes:

\begin{equation}
\begin{split}
P(t) &= P(t_{p0}t_{p1}...t_{pn}) = P(mp_0mp_1...mp_n) \\
& = \Pi_{i=0}^nP(mp_i|mp_0...mp_{i-1})
\end{split}
\end{equation}

To compute Equation 2, there are still two problems to be solved. One needs to collect the sufficient training data, and the other needs to design the parameter training algorithm.

\subsection{Identifying Minimal Phrases In Monolingual Data}
We know from the previous section that identifying the minimal phrases is trivial given the word-aligned parallel sentence pair. Naturally, the corresponding unique partition for the target sentence can be utilized to estimate the language model parameters. However, the bilingual resources are always limited. Therefore, the target part of the bitext is too limited to train a powerful phrase-based language model. The potential is to explore the large-scale monolingual data.

For traditional word-based language model, the target language monolingual data can be used directly after some preprocessing. But, for our language model based on minimal phrases, the first problem is to partition the monolingual sentences into sequence of the minimal phrases. Without source language sentence and the word alignment, identifying the minimal phrases is a difficult problem.

Consider any target language monolingual sentence $t_0t_1...t_n$, our goal is to find the best partition of minimal phrases $t_0..t_i, t_{i+1}..t_k, ..., t_j..t_n$. If we focus on each single word, the task becomes to determine its category: beginning, middle or ending of a minimal phrase {\footnote{We follow the convention of character-based Chinese word segmentation and define the set of categories as \{B, M, E, S\}}}. Consequently, we can formalize the problem as a sequence labelling task. In machine learning, there are many discriminative methods for sequence labelling. We choose the simple but effective perceptron algorithm \cite{collins2002new} to do this job.

The perceptron algorithm maps an input sentence $t \in T$ onto an output minimal phrase sequence $y \in Y$ where $T$ is the set of the target monolingual sentences and $Y$ is the set of all the possible minimal phrase partitions. Given an input sentence $t$, the output $F(t)$ is defined as the highest score among the possible minimal phrase partitions for $t$:

\begin{equation}
F(t) = \argmax_{y \in outputs(t)} \Phi(t,y) \cdot W
\end{equation}

Where $\Phi(t,y)$ is the global feature vector and $W$ denotes the corresponding feature weights vector. What remains is to design the feature templates and construct a training corpus to learn the model parameters $W$.

For the feature templates, we assume that the category of the current word is determined by its surrounding words. Specifically, Table 1 shows all the feature templates we have used in our work. 

\begin{table}\small
\begin{center}
\begin{tabular}{l|l}
\hline \bf Feature Template & \bf Explanation \\ \hline
U00:\%x[-2] & second word preceding current word \\
U01:\%x[-1] & first word preceding current word \\
U02:\%x[0] & current word \\
U03:\%x[1] & next word after current word \\
U04:\%x[2] & second next word after current word \\
U05:\%x[-2]\%x[-1] & combination of left two words \\
U06:\%x[-1]\%x[0] & combination of preceding and current words \\
U07:\%x[0]\%x[1] & combination of current and next words \\
U08:\%x[1]\%x[2] & combination of next two words \\
\end{tabular}
\caption{\label{Table 1} Feature templates for minimal phrase identification.}
\end{center}
\end{table}

For the training corpus, we regard the minimal phrase partitions obtained from {\bf automatically} word-aligned parallel sentences (which are used to train translation model in SMT) as the gold data, so that we can learn the feature weights $W$. During training, the Averaged perceptron is used.

With the trained model, we use it to label the large-scale monolingual data. The resulting large-scale minimal phrase partitions plus those obtained from parallel sentences will serve as the training data to perform parameter estimation.

\section{Parameter Estimation with Deep Learning}
Given the large-scale training data of minimal phrase partitions, we are able to train a phrase-based language model. Following the previous conventions, we  adopt a Markov model of order N-1 to calculate the probability of a minimal phrase sequence:

\begin{equation}
\begin{split}
P(t) &= P(mp_0mp_1...mp_n) \\
& \approx \Pi_{i=0}^nP(mp_i|mp_{i-N+1}...mp_{i-1})
\end{split}
\end{equation}

The standard count-based probability models, such as Kneser-Ney back off \cite{kneser1995improved}, are leveraged to estimate the probability of a word given the preceding N-1 words. It can also be utilized to estimate the probability of a minimal phrase given N-1 previous minimal phrases. We call this model MP-KN.

However, this MP-KN model may encounter the data sparsity problem since the granularity of minimal phrases is bigger than the words and the sequence of minimal phrases is less likely to appear frequently. Furthermore, this MP-KN model cannot take full advantage of similar minimal phrases as they are treated totally different in the MP-KN model. Fortunately, the neural network language models \cite{bengio2003neural,vaswani2013nn,devlin2014fast} are able to give a probability for any N-gram sequence and take advantage of arbitrary large context. Thus, we turn to the deep neural networks to train the phrase-based language model.

\subsection{Neural Network Structure}
Our neural network structure is a feed-forward neural network and it is almost identical to the one described in \cite{vaswani2013nn}. 

As Figure 4 shows, the input vector is the concatenation of N-1 minimal phrase context vectors, in which each minimal phrase is mapped onto a 128-dimensional vector using a shared embedding matrix. Through two 256-dimensional hidden layers with rectified linear activation function ($\phi(x)=max(0,x)$) \cite{gutmann2010noise}, we apply softmax function in output layer to calculate the probability for each minimal phrase in the vocabulary:

\begin{equation}
\begin{split}
P(x) &= \frac{exp(D_{mp}(x))}{Z(x)} \\
Z(x) &= \sum_{mp' \in V}exp(D_{mp'}(x))
\end{split}
\end{equation}

Where $x$ denotes one sample, $D_{mp}(x)$ is the raw output layer score of the observed minimal phrase $mp$ and $Z(x)$ is the normalization term.

\begin{figure}
\centering
\includegraphics[scale=0.5]{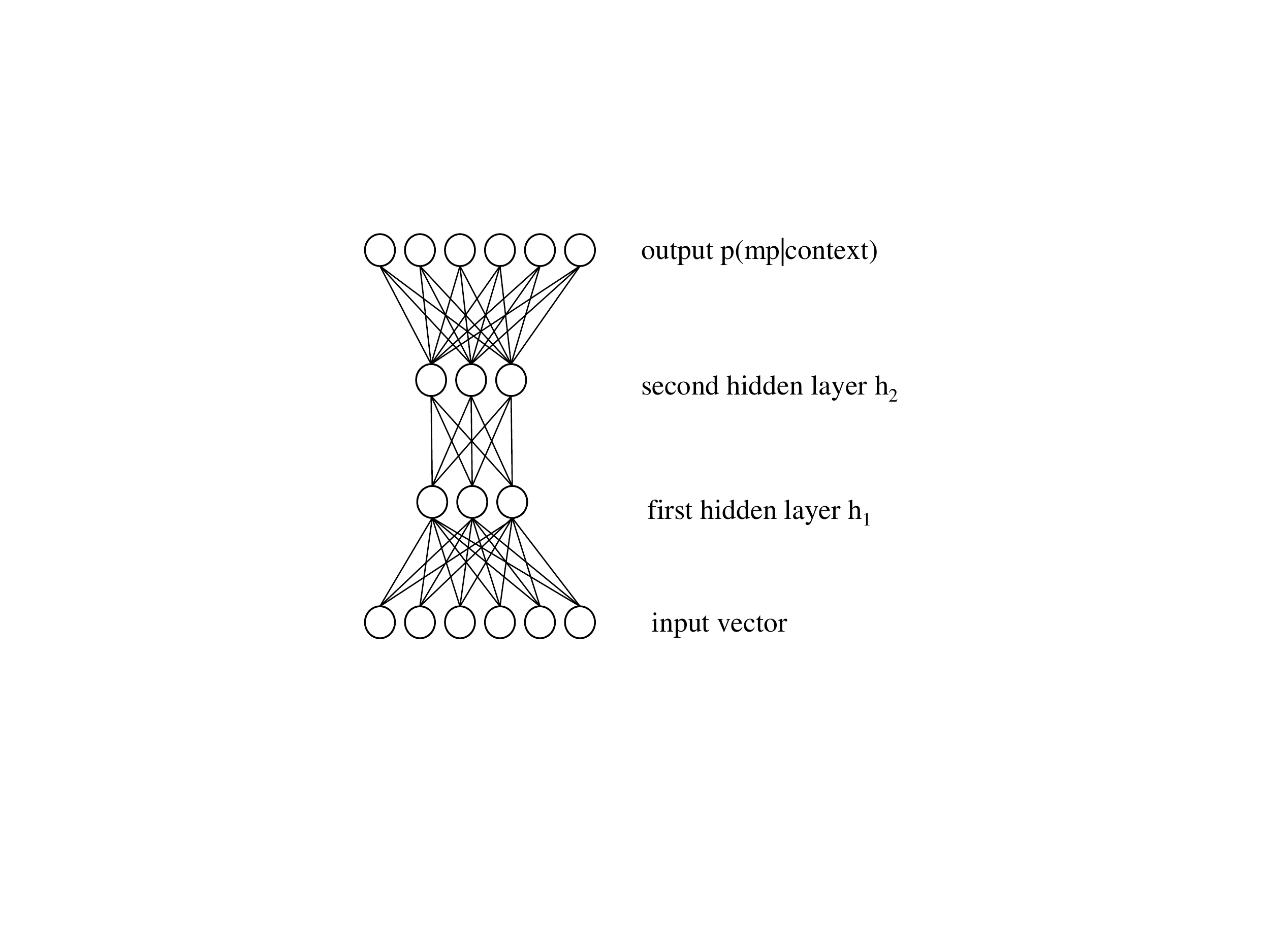}
\caption{Neural network structure for our phrase-based language model.}
\label{Fig.4}
\end{figure}

We should notice that the N-1 minimal phrase context maybe empty or incomplete. The previous methods usually fill the context with {\em OOV} or {\em NULL}, and adopt only one neural network. We believe that the estimated probability would be inaccurate. Instead, we train N-1 neural networks and resort to unigram probability if the context is empty. When calculating the probability of a minimal phrase sequence, we choose the corresponding neural network according to the size of the available context. For example, if $N=5$ (it is used in our experiments), we will build 4 networks $mp\_nn5$, $mp\_nn4$, $mp\_nn3$ and $mp\_nn2$. Suppose we need to compute the probability of the minimal phrase sequence $mp_0mp_1mp_2mp_3mp_4$. $mp\_nn5$, $mp\_nn4$, $mp\_nn3$ and $mp\_nn2$ will be applied to calculate $p(mp_4|mp_0mp_1mp_2mp_3)$, $p(mp_3|mp_0mp_1mp_2)$, $p(mp_2|mp_0mp_1)$ and $p(mp_1|mp_0)$ respectively.

\subsection{Neural Network Training}
Typically, the neural network is optimized using standard back propagation and stochastic gradient ascent algorithm \cite{lecun1998gradient} to maximize the log-likelihood of the training data. However, the softmax layer requires to sum over all the minimal phrases in the vocabulary for each forward computation and it is too time consuming. Recent years witnessed the progress of developing the self-normalized neural networks.

\namecite{vaswani2013nn} adopt the Noisy Contrastive Estimation (NCE) \cite{gutmann2010noise} to avoid the normalization in output layer. In contrast, \namecite{devlin2014fast} explicitly add a constraint for the normalization term ($logZ(x)=0$) in the objective function. In our work, we apply the NCE approach.

The main idea behind NCE is that for each training sample $x$ we choose $k$ noise samples and the network is trained to classify the examples as training sample or noise. The objective function is the conditional likelihood:

\begin{equation}
L = \sum_{i=1}^N(logP(C=1|x_i)+logP(C=0|x_i))
\end{equation}

Here $C=0$ says $x_i$ is noise rather than a training sample.

\section{Experiments}
\subsection{Translation System}
We have implemented a phrase-based translation system with a maximum entropy based reordering model using the bracketing transduction grammars \cite{wu1997itg,xiong2006maxent,zhang2013unified}. In this translation system, the phrasal translation rule $A \rightarrow (x,y)$ converts a source language phrase into a target language phrase, and forms a {\textbf{block}}. The monotone merging rule $A \rightarrow [A^l, A^r]$ combines the two consecutive blocks into a bigger block by concatenating two partial target translation candidates in order while the swap rule $A \rightarrow \langle A^l, A^r \rangle$ swaps the two partial target candidates. Obviously, the phrase-based language model could play an important role in determining whether the result translation (sequence of phrases) are fluent or not.

\subsection{Data Preparation}
The evaluation is conducted on Chinese-to-English translation. The bilingual training data from LDC contains about 2.06 million sentence pairs with 27.7M Chinese words and 31.9M English words. NIST MT03 is used as the tuning data. MT05, MT06 and MT08 (news data) are used as the test data. Case-insensitive BLEU is employed as the evaluation metric. The statistical significance test is performed with the pairwise re-sampling approach \cite{koehn2004sigtest}.

As the language model is the focus of this work, we investigate four language models. The target side raw data includes the English part of the bilingual data and the monolingual Xinhua portion of the English Gigaword. The whole target language data contains about 300 million words. Four different language models are detailed as follows:
\begin{itemize}
\item {\textbf{W-KN}}: It is the conventional word-based language model using Kneser-Ney count smoothing.
\item {\textbf{W-NN}}: It is the word-based neural language model first proposed by \namecite{bengio2003neural} and successfully applied to machine translation by \namecite{vaswani2013nn} and \namecite{devlin2014fast}. They adopt the hierarchical phrase-based model \cite{chiang2007hierarchical} as their baseline while we employ the BTG-based model to be our baseline. In W-NN, we keep the top 160K frequent words in the vocabulary.
\item {\textbf{MP-KN}}: It is the language model using minimal phrases as basic units and trained with Kneser-Ney count smoothing. The raw training data is the same as that of W-KN. Then, the English part of the bilingual data is partitioned into minimal phrases using minimal phrase definition and the Xinhua portion of Gigaword is partitioned into minimal phrases using the sequence labelling algorithm.
\item {\textbf{MP-NN}}: It is the neural language model with minimal phrases serving as basic units using the feed forward neural network introduced in the previous section. The training data is the same as that of MP-KN, but we retain only top 160K frequent minimal phrases in the vocabulary.
\end{itemize}

The different language models will be integrated into the log-linear translation model as the additional features.

\subsection{Experimental Results on Minimal Phrase Partition}
Before detailing the translation performance, we first report the performance of minimal phrase partition in the large-scale monolingual data. We perform word alignment for the Chinese and English reference sentences on NIST MT03. Based on the word alignment constraints, we obtain the English-side minimal phrase partitions which are employed as the gold test data. We apply the trained perceptron algorithm to partition the English reference sentence of NIST MT03. We compare this result with the gold test data. 

The precision, recall and F1-score are \textbf{0.83}, \textbf{0.873} and \textbf{0.851} respectively. It demonstrates that the performance of the minimal phrase partition for the monolingual data is quite good and the minimal phrase partitions on the large-scale monolingual data are very reliable to be used for training the phrase-based language model.

\subsection{Experimental Results on Translation Quality}

\begin{table}\small
\begin{center}
\begin{tabular}{l|c|c|c|c}
\hline \bf Method (Perplexity)  & \bf MT03 & \bf MT05 & \bf MT06 & \bf MT08 \\ \hline
\hline
W-KN  (107.39) & 35.81 & 34.69 & 33.83 & 27.17 \\
\hline
W-NN  (130.12) & 34.73 & 33.62 & 32.75 & 26.54 \\
\hline
MP-KN  (89.95) & 34.39 & 33.26 & 32.51 & 25.65 \\
\hline
MP-NN  (70.55) & 33.65 & 32.83 & 31.96 & 25.21 \\
\hline
\hline
W-KN-NN & {\bf 36.40} & {\bf 35.45} & {\bf 34.58} & {\bf 27.87} \\
\hline
W-KN+MP-KN & 36.26 & {\bf 35.36} & {\bf 34.45} & 25.54 \\
\hline
W-KN+MP-KN-NN & {\bf 36.83} & {\bf 35.87} & {\bf 35.30} & {\bf 28.40} \\
\hline
W-KN-NN+MP-KN-NN & {\bf 36.95} & {\bf 36.13} & {\bf 35.56} & {\bf 28.92} \\
\hline
\end{tabular}
\caption{\label{Table 2} Translation performance of different language model settings. \textbf{Bold} numbers denote that the model significantly outperforms the baseline W-KN with $p < 0.01$.}
\end{center}
\end{table}

To have a comprehensive understanding about how different language models influence the translation quality, we conduct and compare eight language model settings. Table 2 reports the detailed results. 

For the first four lines, we just use one language model in the translation system. We plan to figure out the following two questions: 1, can the neural language model substitute the Kneser-Ney count-based language model? 2, can the phrase-based language model replace the word-based language model in SMT? Comparing W-NN with W-KN in Table 2, we can find that applying only the word-based neural language model cannot perform as well as the word-based Kneser-Ney language model. The similar phenomenon exits for the phrase-based language model (MP-KN vs. MP-NN). In theory, DNN can alleviate the data sparsity problem. However, due to the strict vocabulary size constraint, it cannot substitute the Kneser-Ney language model currently.

About the second question, we are unfortunate to see that the phrase-based language model cannot outperform the word-based language model when comparing MP-KN with W-KN. We find the reason after analysing the decoding process. Since the minimal phrase vocabulary is much larger than word vocabulary (4 vs. 1.7 in million), the n-gram hit rate of the phrase-based language model is much lower than that of word-based language model during decoding. Table 3 gives the statistics on the test sets. Since we apply 5-gram model in both word- and phrase-based language models, the 5-gram hit rate during decoding is a key factor that influences the translation quality. As shown in Table 3 that the hit rate of the phrase-based language model is lower than that of the word-based language model by approximate 10 percent. Therefore, the phrase-based language model has to back off to the lower order model more frequently. We also trained MP-KN with only minimal phrase partitions from bilingual corpus. It achieves the performance of 32.23, 31.09, 30.17 and 23.48 BLEU on MT03, MT04, MT05 and MT06 respectively. Obviously, excluding the large-scale monolingual data degrades the translation quality dramatically.

We also calculate the perplexity of the four language models on the English references of the test sets. As shown in brackets of Table 2, the perplexity has little relationship with the translation quality. We can see that phrase-based language model has the smaller perplexity. However, we believe that they are not comparable as their basic units are different. We also notice that, for the phrase-level, the neural language model has a lower perplexity. But, for the word-level, the neural language model has a higher perplexity. We will further study this phenomenon in our future work.

\begin{table}\small
\begin{center}
\begin{tabular}{l|c|c|c}
\hline \bf Language Model  & \bf MT05 & \bf MT06 & \bf MT08 \\ \hline
W-KN   & 0.2544 & 0.2917 & 0.2347 \\
\hline
MP-KN  & 0.1573 & 0.1911 & 0.1443 \\
\hline
\end{tabular}
\caption{\label{Table 3} 5-gram hit rates for test sets during decoding.}
\end{center}
\end{table}

Although the phrase-based language model cannot surpass the word-based language model when used independently, both of them should be indispensable in measuring the quality of the translation output. To prove this, we incorporate multiple language models into the translation system. The last four lines in Table 2 show the results.

Following \namecite{vaswani2013nn}, we first test the word-based neural language model. Based on the word-based Kneser-Ney language model, the neural language model significantly improves the translation performance (W-KN-NN) with the largest gains 0.76 BLEU on MT05. It is in line with the conclusions in \cite{vaswani2013nn}.

At the basis of word-based Kneser-Ney model, we integrate a Kneser-Ney based phrase language model. The translation results (W-KN+MP-KN in Table 2) demonstrate that the phrase-based language model is much beneficial to improve the translation performance. It outperforms W-KN significantly on MT05 and MT06.

When we further incorporate a neural phrase language model, the translation quality can be upgraded dramatically (W-KN+MP-KN-NN). It obtains significant gains (more than 1.0 BLEU score) over W-KN on all the test sets and the biggest improvement can be up to 1.47 BLEU score on MT06. Moreover, this system even significantly outperform the neural network augmented word-based language model W-KN-NN. It indicates that the neural network can explore deep information (such as syntactic and semantic similarities) of the phrase-based language model.

We are fortunate to see in last line of Table 2 that the improvements of the four language models are additive. It gets the best performance on MT08, outperforming W-KN by 1.75 BLEU score. It can achieve an improvement of more than 1.0 BLEU score over the strong system W-KN-NN. It demonstrates that our proposed phrase-based language model is very helpful in improving the translation quality.

\section{Related Work}
In statistical machine translation, few research work has done well in designing a phrase-based language model.

Many researchers attempt to go beyond the word-based language model and augment the translation system with syntax-based language models. \namecite{charniak2003syntax} design a CFG-based syntax language model for translation output reranking. \namecite{shen2008new} propose a dependency language model for the hierarchical phrase-based system \cite{chiang2007hierarchical}). \namecite{post2009language}, \namecite{xiao2011language} and \namecite{zhang2013syntax} propose a tree substitution grammar based syntax language model for the string-to-tree translation model. However, these syntax-based language models much increase the decoding time and they are very difficult to be integrated into the phrase-based translation systems which just generate translation outputs phrase by phrase.

\namecite{baisa2011chunk} gives a proposal for the chunk-based language model (including phrase-based) in machine translation but does not give a solution. Recently, \namecite{xu2015phrase} present an approach for phrase-based language model in statistical machine translation. Their approach considers any word sequence to be a phrase. It leads to huge phrase vocabulary and severe data sparsity. As a result, the conditional probability between phrases is very difficult to be estimated. They report slight improvements on a small IWSLT data set. In contrast, we give an exact and reasonable definition about minimal phrase, and we also propose a sequence labelling algorithm to  partition the large-scale monolingual data. Furthermore, we adopt the deep neural network to better estimate the conditional probability between minimal phrases. Finally, we obtain significant improvements on the large-scale NIST data set.

The most relevant work to ours is the bilingual n-gram translation model \cite{marino2006n,crego2010improving,durrani2013model,zhang2013beyond,hu2014minimum}. Their Markov model which generates translation by arranging sequence of tuples is very similar to an n-gram language model. The tuple can be any bilingual phrase pair at the early time \cite{marino2006n,crego2010improving}. Recently, tuples become the minimal translation units (MTU) which are the smallest bilingual phrases satisfying the word alignment. \namecite{durrani2013model} and \namecite{zhang2013beyond} perform translation by compositing the MTUs with a Markov model. \namecite{hu2014minimum} apply a recurrent neural network to address the sparsity problem of MTUs. 

Generally speaking, the above MTU-based model can be considered as a bilingual version of our minimal phrase based language model. However, our method is quite different from theirs. First, we focus on language models while they consider translation models. Second, our minimal phrase cannot be empty while they allow empty. Third, their MTUs contain both source- and target-side phrase, and it leads to much more serious sparseness problem than our model. Fourth, the MTUs are inherent bilingual and cannot make use of the monolingual data. In contrast, our phrase-based language model can take full advantage of the large-scale monolingual data.

\section{Conclusion and Future Work}
In this paper, we have presented a novel phrase-based language model for statistical machine translation. We first gave the definition of the minimal phrases. Then, to make full use of the large-scale monolingual data for phrase-based language model training, we developed a sequence labelling algorithm to partition the monolingual data into minimal phrases. Finally, we designed a deep neural network to better learn the parameters of the phrase-based language model. The extensive experiments on Chinese-to-English translation demonstrated that the proposed phrase-based language model significantly improved the translation performance.

In the future work, we plan to explore our phrase-based language model in two directions. For one thing, we are going to further address the sparsity problem of the minimal phrases. For another thing, we will incorporate our phrase-based language model into other translation models, such as the MTU-based translation model and the hierarchical phrase-based translation model.

\bibliographystyle{aaai}
\bibliography{bwblm2015}

\end{document}